\documentclass[nonacm, sigconf,natbib=true]{acmart} 
\AtBeginDocument{%
  }

\usepackage{booktabs}
\usepackage{array}
\usepackage{graphicx}
\usepackage{diagbox}
\usepackage{tabularx}
\usepackage{multirow}
\usepackage{makecell}
\usepackage{colortbl}
\usepackage{listings}
\usepackage{subcaption}
\usepackage{pgfplots}
\pgfplotsset{compat=1.17} 
\usepackage{pgfplotstable} 
\usepackage{amsmath} 
\usepackage{amsfonts} 
\usepackage{tikz}
\usepackage{lipsum}
\usepackage{tcolorbox}
\usepackage[normalem]{ulem}

\AtBeginDocument{\bibpunct{(}{)}{;}{a}{,}{,} \setlength{\bibsep}{0pt} \setlength{\itemsep}{0pt plus 0.3ex} \setlength{\parsep}{0pt plus 0.3ex}}
\AtBeginDocument{\setcitestyle{authoryear,open={(},close={)},maxnames=2,minnames=1}}

\PassOptionsToPackage{colorlinks=true,linkcolor=blue,citecolor=blue,unicode=true}{hyperref}

\usepackage[hang,flushmargin]{footmisc}
\newcommand\blfootnote[1]{
    \begingroup
    \renewcommand\thefootnote{}\footnote{#1}
    \addtocounter{footnote}{-1}
    \endgroup
}

\definecolor{bestcolor}{RGB}{144,224,239}
\newcommand{\bestcell}[1]{\cellcolor{bestcolor}\textbf{#1}}

\definecolor{secondcolor}{RGB}{202,240,248}
\newcommand{\secondcell}[1]{\cellcolor{secondcolor}\textbf{#1}}

\usepackage{rotating}
\usepackage{svg}

\usepackage{enumitem}
\usepackage{pifont}

\usepackage{float}

\newcommand{\benchname}{GFMBench}
\newcommand{\modelname}{LangGFM}

\begin{document}

\title[\benchname~\& \modelname]{\modelname: A Large Language Model Alone Can be a Powerful Graph Foundation Model}

\author{
 \textbf{Tianqianjin Lin\textsuperscript{1,2,$\bigstar$}}
 \textbf{Pengwei Yan\textsuperscript{1,2,$\bigstar$}}
 \textbf{Kaisong Song\textsuperscript{2}}
 \textbf{Zhuoren Jiang\textsuperscript{1,$\spadesuit$}}
 \textbf{Yangyang Kang\textsuperscript{2,$\spadesuit$}}
 \\
 \textbf{Jun Lin\textsuperscript{2}}
 \textbf{Weikang Yuan\textsuperscript{1,2}}
 \textbf{Junjie Cao\textsuperscript{2}}
 \textbf{Changlong Sun\textsuperscript{2}}
 \textbf{Xiaozhong Liu\textsuperscript{3}}
\\
 \textsuperscript{1}Zhejiang University\;  \textsuperscript{2}Alibaba Group  \;\textsuperscript{3}Worcester Polytechnic Institute
\\
\texttt{
\small{
\{lintqj, yanpw, jiangzhuoren, yuanwk\}@zju.edu.cn, \{kaisong.sks, yangyang.kangyy,}
}
\\
\texttt{
\small{linjun.lj, junjie.junjiecao\}@alibaba-inc.com, }
\small{changlong.scl@taobao.com, xliu14@wpi.edu}
}
}
\renewcommand{\shortauthors}{Lin and Yan et al.}



\begin{abstract}
Graph foundation models (GFMs) have recently gained significant attention. However, the unique data processing and evaluation setups employed by different studies hinder a deeper understanding of their progress. Additionally, current research tends to focus on specific subsets of graph learning tasks, such as structural tasks, node-level tasks, or classification tasks. As a result, they often incorporate specialized modules tailored to particular task types, losing their applicability to other graph learning tasks and contradicting the original intent of foundation models to be universal. 
Therefore, to enhance consistency, coverage, and diversity across domains, tasks, and research interests within the graph learning community in the evaluation of GFMs, we propose \benchname—a systematic and comprehensive benchmark comprising 26 datasets. Moreover, we introduce \modelname, a novel GFM that relies entirely on large language models. By revisiting and exploring the effective graph textualization principles, as well as repurposing successful techniques from graph augmentation and graph self-supervised learning within the language space, \modelname~achieves performance on par with or exceeding the state of the art across \benchname, which can offer us new perspectives, experiences, and baselines to  drive forward the evolution of GFMs. 
\end{abstract}




\settopmatter{printacmref=false}
\maketitle
\blfootnote{\noindent $\bigstar$ Equal Contribution $\spadesuit$ Co-corresponding}
\section{Introduction}
\begin{figure*}[ht]
 \centering
 \includegraphics[width=\textwidth]{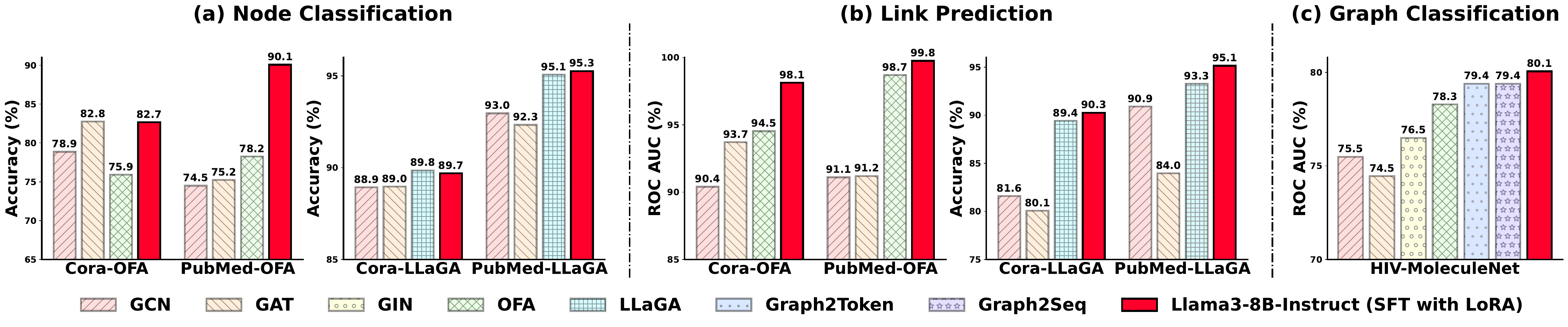}
 \caption{Comparison among GNNs and GFMs across node-, edge- and graph-level tasks. Results of OFA~\citep{iclr2024ofa}, LLaGA~\cite{icml2024llaga}, Graph2Seq~\cite{icml2024graph2seq}, and Graph2Token~\cite{icml2024graphtoken} are all sourced from the best-reported results in their works. 
 It's observed: (1) Different works claim performance within significantly different intervals on the same-name dataset, e.g., the results of OFA and LLaGA in node classification, or they use different metrics, e.g., the metrics employed by OFA and LLaGA in link prediction;
 (2) Converting graphs into texts and performing instruction tuning on LLaMA3-8B-Instruct can consistently outperform the current state-of-the-art. 
 These findings motivate us to develop a benchmark to facilitate fair comparison among GFMs and to provide a simple yet powerful GFM for future development.}
 \label{fig:motivation}
\end{figure*}

Foundation models are the dominant paradigm in artificial intelligence for the 2020s~\citep{bommasani2022opportunitiesrisksfoundationmodels}. They are characterized by their ability to be effectively trained on massive and diverse datasets in a unified manner~\citep{moor2023foundation, zhou2023comprehensivesurveypretrainedfoundation} and are expected to exhibit superior performance in various downstream tasks~\citep{AI_Foundation_Model_Transparency_Act_2023}. 
Large language models (LLMs) exemplify this paradigm, achieving outstanding performance across nearly all text-related tasks. 
Their success is largely attributed to the unification of both input and output processes, leveraging a closed-set vocabulary and tokenizer to transform diverse textual data into a common embedding space, and reframing various tasks—such as classification, extraction, and summarization into text-to-text generation. This enables LLMs to seamlessly handle diverse tasks across different domains concurrently.

In contrast, the development of foundation models in the graph domain has been notably slower. Currently, there is no counterpart to the closed-set ‘vocabulary and tokenizer’ for graph inputs, nor is there a ‘text-to-text’ framework for unifying graph tasks~\citep{mao_PositionGraphFoundation_2024, iclr2024ofa, icml2024llaga}. 
One key challenge on the input side stems from the intrinsic heterogeneity of nodes and edges in graph data. Unlike text, which can be represented with a finite vocabulary, graphs from different domains exhibit varying types of nodes and edges, along with distinct attribute dimensions for these nodes and edges. As a result, current graph models require the prior specification of these types and dimensions for specific datasets, making it difficult to transfer models across different datasets at the code level. 
On the output side, the challenge lies in the varying scales of graph tasks, which are commonly classified into node-level, edge-level, and graph-level tasks~\citep{ogb}. Each of these tasks requires different prediction pipelines. For instance, pooling layers are essential for graph-level tasks but are unnecessary for node-level predictions.  
Therefore, when current graph models are applied across tasks of different levels, they often need to learn task-oriented modules from scratch.

To drive the progress of graph foundational models (GFMs), several recent efforts have emerged. On the input side, these methods capitalize on natural language descriptions of node and edge types, as well as features, which are subsequently processed by language models to generate fixed-size vector representations~\citep{iclr2024ofa}. This approach avoids the need to pre-define node and edge types and dimensions, enhancing flexibility across datasets. 
On the output side, multiple technical directions are adopted. One option uses graph neural networks to encode the graph, followed by graph prompting techniques to facilitate cross-task learning~\citep{iclr2024ofa}. Another approach, drawing inspiration from advancements in multimodal LLMs, trains adapters to map graph vectors into specialized tokens. These tokens are then fed into LLMs, enabling predictions via in-context learning~\citep{icml2024llaga}. 

However, two significant concerns remain unaddressed and may hinder the development of GFMs. \textbf{\textit{Firstly}}, our understanding of how current works actually make progress has been far limited by the different data processing and evaluation settings adopted by them. For instance, we can consider the cases of OFA~\citep{iclr2024ofa} and LLaGA~\citep{icml2024llaga}. As illustrated in Figure~\ref{fig:motivation}a, both OFA and LLaGA utilize the Cora and PubMed datasets for the node classification task. Notably, OFA achieves an accuracy of less than 80\% on these datasets, while LLaGA reaches approximately 90\%. Actually, this disparity is largely caused by the different label rates: OFA operates under a low-label-rate setting, whereas LLaGA operates under a high-label-rate setting. 
Additionally, OFA and LLaGA employ different evaluation metrics for the same dataset and task. As shown in Figure~\ref{fig:motivation}b, OFA uses the ROC AUC metric, while LLaGA relies on accuracy for link prediction. 
These discrepancies make it difficult to compare the capacity of OFA and LLaGA objectively. Referring to the development of LLMs, the establishment of a standard benchmark, such as MMLU, is essential for fostering meaningful evaluations across different models.

\begin{figure}[ht]
 \centering
 \includegraphics[width=\linewidth, trim=0.5cm 17cm 11cm 11cm, clip]{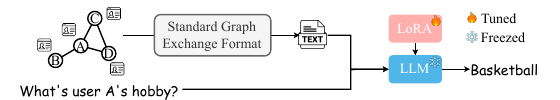}
 \caption{An illustration of instruction tuning for an LLM to perform graph tasks.}
 \label{fig:langgfm}
\end{figure}

\textbf{\textit{Secondly}}, current research tends to concentrate on limited subsets of graph learning tasks. For example, OFA targets classification tasks, while LLaGA focuses on node-level and edge-level tasks. As a result, they often incorporate modules tailored to particular task types, limiting their applicability to other graph learning tasks and thus contradicting the original intent of foundation models to be universal. 
Thus, a question naturally arises: \textbf{\textit{At the cost of sacrificing versatility, have they achieved unparalleled performance on certain subsets of tasks?}} In response to this question, we consider a fully unified approach (as shown in Figure~\ref{fig:langgfm})—representing graphs using plain text via standard graph exchange format like GraphML~\citep{graphml_format} and directly fine-tuning an LLM with the task instruction. 
Surprisingly, this approach surpassed the current state-of-the-art models in almost all task settings as shown in Figure~\ref{fig:motivation}. Even in highly specialized domains such as molecular graph property prediction, this approach outperforms domain-specific foundation models: as shown in Figure~\ref{fig:motivation}c, this solution achieved an ROC AUC score of 80.1, which is higher than the best ROC AUC reported in the latest works like Graph2Token~\citep{icml2024graphtoken} and Graph2Seq~\citep{icml2024graph2seq}. 
  
These exciting experimental results inspire us to propose a provocative question: \textbf{\textit{Is there a genuine necessity for specialized modules, such as graph neural networks and graph transformers, in graph processing?}} We argue that such modules may be superfluous for two primary reasons.
First, the expressive capacity of contemporary graph models is limited~\citep{jure2022expressive}, resulting in unavoidable information loss when encoding graph structures with these approaches. To build GFMs, it is essential to reconsider reliance on these traditional architectures. Second, current development of GFMs has been largely influenced by the success of multimodal LLMs, which adeptly process non-textual data through specialized encoders and subsequently integrate them into the LLMs via projection. However, it is critical to recognize that the graph is essentially a data structure and/or mathematical language rather than a modality. 
We can only represent and transmit images or audio using digital signals, but we do often describe and transmit graphs using language, at least in the form of text files like CSV or JSON. For instance, the first digital image emerged in 1957, yet research on graphs (e.g., social network analysis) can be traced back to the 1930s. 

In light of these insights, this work contributes three significant advancements that propel the development of GFMs: 
\begin{itemize}
    \item We introduce \benchname, a systematic and comprehensive benchmark encompassing 26 datasets. This benchmark is meticulously crafted to ensure consistency on the evaluation pipeline, diversity in graph domains, and broad coverage across tasks and research interests within the graph learning community; 
    \item We present \modelname, a GFM entirely built on the LLM. \modelname~distinguishes itself from prior works through: reflective exploration and formulation of effective principles for graph learning textualization, innovative strategies for graph data augmentation, and graph self-supervised learning within the language space, and a fully joint learning attempt that covers all domains and tasks;
    \item We perform extensive evaluations of \modelname~on \benchname, as well as an investigation into its zero-shot learning capabilities on datasets outside of \benchname. \modelname~achieves state-of-the-art performance and robustness compared to existing methodologies, establishing a new benchmark for GFMs and providing invaluable insights for future research and development in this domain.
\end{itemize}

In the following sections, we first provide a review of the existing GFMs with their strengths and weaknesses and analyze the limitations of the datasets or benchmarks used in existing work in Section~\ref{sec:related_work}. Then, we introduce \benchname~in detail in Section~\ref{sec:benchmark} and illustrate \modelname~step by step in Section~\ref{sec:method}. Finally, we report extensive experimental results about \modelname~in Section~\ref{sec:experiment}.

\section{Related Works}\label{sec:related_work}

GFM seeks to construct graph models that effectively harness the training from extensive and diverse datasets, thereby demonstrating superior applicability and performance across a wide range of tasks and domains~\citep{mao_PositionGraphFoundation_2024,liu_GraphFoundationModels_2023}. 
  
Earlier efforts focused on leveraging self-supervised learning~\cite{hu2020gpt,qiu2020gcc,hou2022graphmae,zhu2021graph} to pre-train GNNs on large graph data, then finetune the GNNs or {design graph prompting techniques for downstram tasks}~\citep{kdd2023allinone}. However, due to the inherent limitation of the GNN architecture, the applicability of these methods remains confined to the same domain, and these methods require training of new parameters for different downstream tasks.

Recently, the success of LLMs and multimodal LLMs has advanced the development of GFMs. 

First, LLMs can unify the input inconsistency across different graphs. \citet{iclr2024ofa} describes the types and attributes of nodes and edges using natural text, then uses LLMs to encode these texts into fixed-size semantic vectors in a shared space, which are treated as new features for the nodes and edges. In this way, graphs from various domains can share the same GNN encoder. However, they are still limited to the same type of tasks since the GNN backbone is retained as the predictor in the overall framework. For instance, OFA~\citep{iclr2024ofa} and ZeroG~\citep{zerog} are restricted to classification tasks. 

Second, similar to multimodal LLMs, LLMs can serve as a unified predictor. The representations of nodes, edges, or graphs can be projected into the embedding space of the LLM, allowing the LLM to predict various tasks through prompting directly~\citep{icml2024llaga,tang_GraphGPTGraphInstruction_2024}. However, these approaches can suffer from loss of graph structural information, as the LLM can not see the original graph structure for prediction. As a result, these works mainly focus on tasks like node classification, node description, and link prediction, which do not heavily depend on graph structure.

At the same time, some studies~\citep{wang2024can,guo2023gpt4graph} from the NLP community have explored the graph understanding and reasoning capabilities of LLMs. However, these works are limited to evaluations and prompt engineering. Inspired by the instruction tuning, recent works have further enhanced the graph reasoning ability of LLMs through instruction tuning~\citep{wang_InstructGraphBoostingLarge_2024,chen2024graphwiz}. However, these works focus solely on improving performance through LLM-finetuning techniques such as direct preference optimization, while overlooking knowledge and experience from the graph learning community, such as graph data augmentation and graph self-supervised learning. Additionally, the tasks these studies investigate differ markedly from those traditionally prioritized by graph mining works. For example, they do not tackle graph-level tasks like molecular property prediction. 

In summary, the current GFM has only been developed to generalize in a limited set of domains and tasks. 
Following \citet{mao_PositionGraphFoundation_2024}, we refer to these related works as primitive GFM. 

One potential factor affecting the development of GFM is that the current benchmarks, which are known as the foundation for model development, lack systematic design and do not align with the expectations for GFM. 
  
(1) The isolation between graph structure learning tasks and real-world applications. Models that excel in understanding graph structures but cannot solve real-world problems are of limited value, and models that perform well in real-world tasks but fail to understand basic graph structures lack trustworthiness. However, most existing benchmarks focus only on one of these aspects. For instance, NLGraph~\cite{wang2024can} only examines basic graph structure tasks like the shortest path problem. \citet{chen_ExploringPotentialLarge_2024} only investigate tasks like academic paper classification. The negative impact of this phenomenon is that subsequent research has become fragmented, preventing a collective effort to create GFM.
  
(2) Insufficient coverage in real-world graphs and applications. 
In terms of graphs, for instance, GLBench~\cite{li2024glbench} and TS-GFM~\cite{chen2024text} focus on text-rich graphs like citation networks, but overlooking text-irrelevant graphs, e.g., brain networks. Additionally, while graphs with dynamic~\citep{zhang2024llm4dyg} and heterogeneous~\citep{tang2024higpt} properties are common and are worthy to be well-evaluated, they are excluded by most works.
From the perspective of applications, for example, OFA and TS-GFM only focus on classification tasks, ignoring important regression tasks like molecular free energy prediction. Moreover, few works investigate graph-related open-ended tasks, e.g., graph description. To establish an ideal GFM benchmark to verify and advance the versatility of GFM, it is essential to consider various graph characteristics and applications. 

\section{The proposed benchmark}\label{sec:benchmark}

Our benchmark provides a comprehensive evaluation of graph understanding by encompassing a systematical design of a diverse array of tasks, which can be divided into structural understanding and semantic learning tasks. 
In terms of structural tasks, the capability of LLMs to understand the topological structures of synthetic graphs in different scales is evaluated. We incorporate entity-based, path-based, and structure-based challenges, organized according to task complexity. 
For semantic learning tasks, we aim to testify how GFMs perform on reasoning with various semantic-featured graphs. Semantic-featured graphs refer to graphs that have realistic meanings for entities or relations within the graph. Specifically, we consider factors such as graph domain, data heterogeneity, and data dynamism in terms of graph diversity, and construct varying levels and types of graph tasks that adapt to the new GFMs trend. 

\begin{table*}[]
\centering
\caption{Graph structure understanding tasks of \benchname.}
\setlength{\tabcolsep}{5pt}
\label{tab:benchmark-structure}
\begin{tabular}[h]{ccc}
\toprule
Task   Level                      & Task                 & Time Complexity               \\
\midrule
\multirow{3}{*}{Entity-Based}    & Graph Size (Node \& Edge)           & $O(|V|+|E|)$                    \\
                                 & Attribute Retrieval (Node \& Edge)   & $O(|V|+|E|)$                    \\
                                 & Degree Count         & $O(|E|) $                       \\
\midrule
\multirow{3}{*}{Path-Based}      & Shortest Path        & $O(|E|+|V|\log|V|)$              \\
                                 & Maximum Triangle Sum & $O(|V|^{3})$         \\
                                 & Hamilton Path        & NP-Complete                   \\
\midrule
\multirow{3}{*}{Structure-Based} & Sugraph Matching     & NP-Complete                   \\
                                 & Graph   Structure    & NP-Complete                   \\
                                 & Graph   Automorphism & NP-Complete                   \\
\bottomrule
\end{tabular}
\end{table*}
\begin{table*}[!ht]
    \caption{Graph semantic understanding tasks of \benchname.}
    \label{tab: semantic}
    \centering
    \begin{tabular}{ccccccc}
    \toprule
    Dataset                           & Domain     & Task Level & Text-driven & Dynamic & Graph Type    & Task Type \\ 
    \toprule
    Ogbn-Arxiv                        & Academic   & Node       & Yes         & Yes     & Homogeneous   & Multi-class Classification \\ 
    WikiCS                            & Web        & Node       & Yes         & No     & Homogeneous   & Multi-class Classification \\ 
    Twitch                            & Social     & Node       & No          & Yes     & Homogeneous   & Binary Classification \\ 
    USA Airport                       & Traffic    & Node       & No          & No     & Homogeneous   & Ordinal Regression \\ 
    AMiner                            & Academic   & Node       & Yes         & No     & Heterogeneous & Multi-class Classification \\ 
    \midrule
    FB15K237                          & Knowledge  & Link       & Yes         & No     & Heterogeneous   & Multi-class Classification \\ 
    Ogbl-Vessel                       & Brain      & Link       & No          & No     & Homogeneous   & Binary Classification \\ 
    MovieLens                         & Social     & Link       & Yes         & Yes     & Heterogeneous & Ordinal Regression \\ 
    \midrule
    Fingerprint                       & Vision     & Graph      & No          & No     & Homogeneous   & Multi-class Classification \\ 
    BACE                              & Molecule   & Graph      & No          & No     & Homogeneous   & Binary Classification \\ 
    ESOL                              & Molecule   & Graph      & No          & No     & Homogeneous   & Regression \\ 
    \midrule
    Twitter Friend Circle Description & Social     & Open-ended & No          & No      & Homogeneous & Text Generation\\ 
    Yelp Review Generation            & Social     & Open-ended & Yes         & Yes     & Heterogeneous & Text Generation\\ 
    Molecular Description             & Molecule   & Open-ended & No          & No     & Homogeneous & Text Generation\\ 
    Cypher Query Generation           & Code       & Open-ended & No          & No     & Heterogeneous & Text Generation\\ 
    \bottomrule
    \end{tabular}
\end{table*}

\subsection{Graph Structure Understanding Tasks}
Graphs are structures representing relations among entities, where nodes represent entities and edges represent their connections or interactions. 
A key prerequisite for performing graph machine learning tasks, particularly those GFMs involving LLMs, is a clear understanding of the structure-related concepts in graphs.
Unlike graph neural networks (GNNs) like GCN~\cite{kipf2016semi}, GAT~\cite{velivckovic2017graph}, and GIN~\cite{xu2018how} which are targeted to operate on the unique non-Euclidean data structure of graphs, language models are born to address sequence data, especially text. Thus, whether the LLMs can correctly understand the basic concepts and structures in graph data needs to be testified. 

Inspired by ~\citet{wang2024can} and GraphWiz~\cite{chen2024graphwiz}, we first include graph structure understanding tasks in terms of different difficulty levels and provide benchmark datasets in various sizes to fully check the understanding capability of GFMs for graphs. 
These tasks include \textbf{Graph Size} (for both node and edge sizes), \textbf{Attribute Retrieval} (for both that of nodes and edges), \textbf{Degree Counting, Shortest Path, Maximum Triangle Sum, Hamilton Path, Subgraph Matching, Graph Structure, and Graph Automorphsim}. The summary of graph structure understanding tasks is listed in Table~\ref{tab:benchmark-structure} and detailed information is included in supplementary.
These graph structure understanding tasks are organized according to entity-based, path-based, and graph-based, and the difficulties of tasks are gradually built up. Worth mentioning, we also include graph isomorphism in our benchmark, which is a fundamental problem in graph theory that involves determining whether two graphs are structurally identical. It serves as a basis for understanding more complex graph-related problems and many real-world applications~\cite{graphIso}, such as pattern recognition, network security, and bioinformatics, require solving graph isomorphism problems. Besides, the graph isomorphism problem has been used to testify the expression power of graph neural networks~\cite{li2022expressive,xu2018powerful}.

For the datasets of tasks mentioned above, we utilize a programming aid approach~\cite{chen2024graphwiz} to create random synthetic graphs tailored for each specific task. Every task is linked to a distinctive template designed to reflect the unique properties of graphs, such as whether they are directed or undirected and whether their edges are weighted. To generate the random graphs, we employ the Erdős-Rényi (ER) model, which requires two parameters: the number of nodes \( n \) and the probability \( p \) that an edge exists between any pair of nodes. For each node pair, the generator randomly determines whether to form an edge based on probability \( p \), leading to a graph with an average edge density of \( p \). We used the NetworkX library
to create the random graphs and to solve the graph-related tasks.

\subsection{Graph Semantic Learning Tasks}
Different from structure learning, graph semantic learning tasks are constructed with graphs being connected with specific real-world scenarios, with nodes representing real entities and edges showing specific relations. As a fundamental form of data organization, graphs appearing in social networks, molecule graphs, and citation networks are ubiquitous and valuable in the real world~\cite{wu2020comprehensive}. Considering the various types of real-world graphs and task complexity, we propose a comprehensive graph semantic learning benchmark in terms of \textit{graph domain, graph heterogeneity, graph text-richness, and task level}. 

Firstly, \textbf{Node Classification, Link Prediction, and Graph Classification,} which are three dominant graph learning tasks in traditional graph machine learning~\cite{wu2020comprehensive}, are included. Furthermore, as foundation models are required to couple with tasks with different tasks and adapt to different types of labels~\cite{moor2023foundation}, \textbf{Graph Regression} and \textbf{Open-Ended Graph Understanding} are also included in our benchmark. Inspired by the foundation models like GPT for NLP tasks, especially the generative mode that provides a flexible application interface, we believe that the LLM-based graph foundation models could also have the ability to solve important graph open-ended problems like Graph Q\&A. In terms of the graph domain, we include a wide spectrum of nine scenarios from academic, social media, knowledge graph, biology, chemistry, and so on. What's more, the different graph heterogeneity and text-richness are also considered and included to comprehensively check the graph semantic understanding capability of graph foundation models. All graphs within the datasets are aligned with describing in the natural text according to the background and definition of the graphs, taking both node attributes and edge attributes into account if available. Table~\ref{tab: semantic} shows the detailed introduction for included datasets.

\subsection{Evalution Types and Pipelines}
To sum up, \benchname~ includes 26 datasets, in which 19 classification tasks, 3 regression tasks, and 4 generation tasks. 
For each dataset, we take a random split with \textit{train: valid: test} as \textit{500: 100: 200}, based on the original dataset split. For classification and regression tasks, we take \textit{Accuray} and \textit{RMSE} as metrics to evaluate the performances, and for generation tasks, \textit{ROUGE-L} is deployed to evaluate the consistency of generated content with ground truth.

\section{Methodology}\label{sec:method}

Here, we present \modelname, a GFM fully grounded in the capabilities of LLMs. We begin by outlining the conventional paradigms of graph machine learning, identifying their inherent limitations, and then introducing the concept of GFM alongside the associated challenges. Subsequently, we delve into the foundational ideas that underpin \modelname. We critically examine and articulate effective principles for textualizing graph learning. Moreover, we propose innovative strategies inspired by successful experience in \textit{graph augmentation} and \textit{graph self-supervised learning}, aiming to enhance the graph comprehension and reasoning capabilities of \modelname.

\subsection{Preliminaries}~\label{sec:method_background}
\vspace{-1em}
\subsubsection{Classical Graph Machine Learning and Limitations}
A graph is a data structure used to describe relationships (edges) between objects (nodes). Examples include social networks, citation networks, knowledge graphs, and molecular structures.  
Graph machine learning focuses on predicting properties associated with nodes, edges, and entire graphs. 
In node-level and edge-level tasks, the input typically consists of the ego-graph~\citep{egograph} surrounding the target node or edge within a broader graph~\citep{kdd2023allinone}, e.g., the followers of a Twitter user. For graph-level tasks, the input is usually a complete, independent graph, such as a specific chemical molecule.  
  
Let \(\pi_i\) represent a specific graph machine learning problem, \(\mathcal{G}_{i}\) denote a set of ego-graphs or graphs, and \(\mathcal{Y}_{i}\) be the target label space. \(\pi_i\) refers to learning a mapping function \(f_{\pi_i}\) from \(\mathcal{G}_{i}\) to \(\mathcal{Y}_{i}\): 

\begin{equation}
f_{\pi_i}: \mathcal{G}_{i} \rightarrow \mathcal{Y}_{i},\; f_{\pi_i}(G_{i}^j) = Y_{i}^j,
\end{equation}
where \(G_{i}^j \in \mathcal{G}_{i}\) and \(Y_{i}^j \in \mathcal{Y}_{i}\) are the $j$-th graph and its corresponding label. 
Currently, Graph Neural Networks and Graph Transformers are the leading approaches for modeling \(f_{\pi_i}\)~\citep{GNNGT_survery}. 
Their core concept is to represent nodes by iteratively aggregating information from other nodes within a defined receptive field. Once the node representation is obtained, edge representation is derived by merging endpoints, and graph representation is formed by reading out all the involved nodes. These representations are then transformed into label values using predictors like multi-layer perceptrons (MLPs).  
  
Despite the effectiveness, \(f_{\pi_i}\) typically has a one-to-one correspondence with \(\pi_i\) due to the characteristic of Graph Neural Networks and Graph Transformers. On the input side, the process of representing nodes is tied to the schema of the input graph; on the output side, tasks at different levels have their own specific encoding pipelines, and different label spaces require distinct predictors. Therefore, it is often necessary to reinitialize and train a separate \(f_{\pi_i}\) for each \(\pi_i\). 
This results in at least two limitations: \(f_{\pi_i}\) can not generalize knowledge from diverse graph data, and it has no zero-shot transfer capacity. 

\subsubsection{Graph Foundation Model and Challenge}
In contrast to classical graph machine learning, GFM is conceptualized as a comprehensive and unified model that can effectively handle a wide spectrum of graph machine learning problems at the same time.  
  
Formally, denoting the set of all the graph machine learning problems as $\Pi=\{\pi_i\mid i\in\{1,2,\dots, m\}\}$, the ideal GFM can be expressed as a single function:
\begin{equation}
f_{GFM}: \bigcup_{\pi_i \in \Pi} \mathcal{G}_{i} \rightarrow \bigcup_{\pi_i \in \Pi} \mathcal{Y}_{i}, \; f_{GFM}(G_{i}^j) = Y_{i}^j. 
\end{equation}

From the perspective of model implementation, the key challenge and prerequisite in achieving \( f_{GFM} \) lies in finding standardized spaces to accommodate \(\bigcup_{\pi_i \in \Pi} \mathcal{G}_{i}\) and \(\bigcup_{\pi_i \in \Pi} \mathcal{Y}_{i}\). 

\subsection{Overarching Philosophy of \modelname}
As the saying goes, ``The limits of my language mean the limits of my world''~\citep{wittgenstein2017limits}, it is a potential solution to describe \(\bigcup_{\pi_i \in \Pi} \mathcal{G}_{i}\) and \(\bigcup_{\pi_i \in \Pi} \mathcal{Y}_{i}\) in natural language~\citep{guo2023gpt4graph}. 
By employing a language-based approach to encode graphs and labels, we can facilitate interoperability among heterogeneous datasets and learning tasks. 

\subsubsection{Language as the Standardized Spaces}
Before we elaborate on the details of representing graphs with natural language, we first articulate the high-level idea behind it.
Formally, this philosophy can be expressed as: 
\begin{equation}
    \mathcal{I} = \{I_i^j \mid I_i^j = \mathrm{Lang}_{G}(G_{i}^j, \pi_i),\; G_{i}^j \in \bigcup_{\pi_i \in \Pi} \mathcal{G}_{i} \}
\end{equation}

\begin{equation}
    \mathcal{O} = \{O_i^j \mid O_i^j=\{\mathrm{Lang}_{Y}(Y_{i}^j, \pi_i),\; Y_{i}^j \in \bigcup_{\pi_i \in \Pi} \mathcal{Y}_{i} \}
\end{equation}

where \(\mathrm{Lang}_{G}(\cdot)\) and \(\mathrm{Lang}_{Y}(\cdot)\) are supposed to be methods for transforming an input graph $G_i^j$ and its corresponding label $Y_i^j$ into natural language in the context of a specific learning problem $\pi_i$, respectively. Hence, the function \( f_{GFM} \) can be reformulated as a text-to-text model: 
\begin{equation}
f_{GFM}: \mathcal{I} \rightarrow \mathcal{O},\; f_{GFM}(I_i^j) = O_i^j
\end{equation}

It's widely recognized that LLMs are optimal for text-to-text tasks. Thus, we assume \(f_{GFM}\) is implemented by an LLM, now we briefly overview the generation and training mechanisms.

\subsubsection{Generation Process of LLMs} LLMs employ an autoregressive approach for text generation. Given an input sequence \(X = (x_{1}, x_{2}, ..., x_{t-1})\), the model predicts the next token \(x_t\) as follows:

\begin{equation}
    x_t = \arg\max_{x' \in V} \textrm{P}(x' | x_{1}, x_{2}, \ldots, x_{t-1}; \Theta)
\end{equation}

Here, \(V\) denotes the vocabulary, and \(\textrm{P}(x' | x_{1}, x_{2}, \ldots, x_{t-1}; \Theta)\) is the conditional probability of \(x'\) given the preceding tokens and model parameters \(\Theta\). The token with the highest probability is selected, and this process repeats until a predetermined sequence length is met or a special end token is generated.

\subsubsection{Loss Function of Instruction Tuning} To optimize the LLM-based $f_{GFM}$, instrution tuning is adopted. For a sample \((I, O)\), where \(I \in \mathcal{I}\) and \(O \in \mathcal{O}\), we use a cross-entropy-based loss function: 

\begin{equation}
    \mathcal{L}(I, O) = -\sum_{t=1}^{T} \sum_{v \in V} O_t^v \log(\textrm{P}(O_t = v | I, O_{:t-1}; f_{GFM}))
\end{equation}

In this equation, \(O_t^v\) is the target label at position \(t\), and \(\textrm{P}(O_t = v | I, O_{:t-1}; f_{GFM})\) represents the model’s predicted probability of that label given the input and preceding output. Here, \(T\) is the length of the output sequence, and \(V\) represents the vocabulary. Minimizing \(\mathcal{L}\) improves the model's adaptability to task instructions, enhancing performance on specific tasks. Consequently, the GFM learning problem can be expressed as:

\begin{equation}
    \mathcal{L}_{GFM} = -\sum_{i=1}^{m}\sum_{j=1}^{n_i} \mathcal{L}(I_i^j, O_i^j),
\end{equation}

where \(m\) is the number of learning problems and \(n_i\) is the sample size for the \(i\)-th problem.

\subsection{Textualization of Graph Learning}\label{sec:graph_text_repr}
The only problem that remains in the above is how to define and implement \(\mathrm{Lang}_{G}(\cdot)\) and \(\mathrm{Lang}_{Y}(\cdot)\) and conduct instruction tuning on LLM. We illustrate them here. 

\subsubsection{A Toy Example} In summary, \(\mathrm{Lang}_{G}(\cdot)\) involves representing \(G_i^j\) in text and using natural language to describe the task \(\pi_i\) with specification on target nodes, edges, or the entire graph, including \textit{Graph Description}, \textit{Graph Text} and \textit{Query}. \(\mathrm{Lang}_{Y}(\cdot)\) describes the value \(Y_i^j\) in natural language within the semantic context of the task \(\pi_i\). A toy example on social network is shown below.

\begin{center}\small \begin{tcolorbox}[title={\textcolor{gray!10}{\textbf{Naive Example of Textualization of Graph Learning}}}, colback=lightgray!10, colframe=gray!80, rounded corners=all, width=0.47\textwidth]
    \textbf{Input: $\boldsymbol{I_i^j \leftarrow \mathrm{Lang}_G(G_i^j,\pi_i)}$} \\
    \text{\# \underline{Graph Description}: introducing the graph context.}  \\
    \textcolor{gray}{\textit{This is a social network where nodes are users and edges}} \\
     \text{\textcolor{gray}{\textit{represent following relationships.}}} \\
    \text{\# \underline{Graph Text}:} describing graph $G_i^j$ in natural language. \\
    \text{\textcolor{gray}{\textit{The first user likes basketball and football; the second user}}}\\
    \text{\textcolor{gray}{\textit{is interested in economics...}}} \\
    \text{\textcolor{gray}{\textit{The first user follows the second user...}}}\\ 
    \text{\# \underline{Query}:} \text{describing $\pi_i$ on target node/edge/graph. }\\
    \text{\textcolor{gray}{\textit{What could be the occupation of the second user?}}}  \\
    \textbf{Output: $\boldsymbol{O_i^j \leftarrow \mathrm{Lang}_Y(Y_i^j,\pi_i)}$}\\
    \text{\# \underline{Anwser}:} {Explaining $Y_i^j$ with the task and target.}\\
    \text{\textcolor{gray}{\textit{The profession of the second user could be ...}}}
\end{tcolorbox}\end{center}

\subsubsection{Available Approaches for Graph Text}\label{sec:graph_repr_schemes}

The most important yet challenging part in the above example is \textit{Graph Text}. 
Drawing inspiration from traditional research in social network analysis and graph visualization, we turned our attention to graph-exchange file formats. 
Over the past two decades, extensive efforts have been made to develop flexible, concise, and efficient formats to facilitate graph data exchange and support scientific exploration across a wide range of applications,  
which has led to the creation of nearly one hundred distinct file formats~\cite{arxiv2015graph_exchange}. 
  
In this work, we focus on four prominent formats selected for their widespread adoption and representational expressiveness: \textit{Graph Modelling Language (GML)}~\citep{gml_format}, \textit{Graph Markup Language (GraphML)}~\citep{graphml_format}, \textit{JavaScript Object Notation (JSON)} and \textit{Markdown Table}~\citep{markdown_table}. 


Despite the extensive research on graph-exchange formats, most current works~\citep{wang_InstructGraphBoostingLarge_2024,chen2024graphwiz} have overlooked these efforts, instead focusing on designing custom or intricate text-based representations for graphs. While it is plausible that LLMs have encountered standard formats during pre-training, one could hypothesize that LLMs process graph structures more effectively when presented in these familiar formats, as opposed to custom-designed ones. Unfortunately, existing studies have not thoroughly evaluated the efficacy of their proposed formats against these standard formats. Thus, the necessity of developing new graph representation schemes tailored for LLM-driven graph learning warrants further investigation. In Section~\ref{sec:format_expr}, we perform comparative experiments to examine which text representations LLMs favor when interpreting graphs.

\subsection{Various Formats as Graph Augmentations}\label{sec:method_format_aug}
The different textual representation schemes for graphs remind us of the concept of data augmentation. 
Data augmentation is an effective method for enhancing machine learning model performance since it increases the quantity and diversity of samples and thus improves the model's robustness and generalization ability. 
In the image domain, common techniques include rotation, flipping, cropping, and color transformation; in the text domain, methods such as synonym replacement, random deletion, and back-translation are employed. 
Traditional graph data augmentation techniques enhance model performance through methods like edge masking, node feature perturbation, subgraph sampling, and so on. 
  
Beyond traditional graph data augmentation, the different textual representation approaches inspire us to directly leverage them as a novel data augmentation strategy in the language space. In fact, this aligns well with the core idea of data augmentation—maximizing the variation of input features while preserving semantic equivalence. Obviously, two different format textual representations of a graph exactly describe the same thing, but they possess very distinct natural language characteristics, including sequence length, organizational logic, and so on. Moreover, in the case of in-context learning, recent research reveals that different formats can exhibit significant performance differences across various tasks~\citep{guo2023gpt4graph,fatemi_TalkGraphEncoding_2023}, which indicates bias exists. Intuitively, when LLMs are required to answer the same question about different format representations of the same graph object, they can develop the ability to understand the graph object itself, independent of such format biases. 

Our experimental results in Section~\ref{sec:format_aug} demonstrate the strong effectiveness of this strategy. 

\subsection{Graph Self-supervised Instructions}\label{sec:method_ssl}
Graph self-supervised learning has proven to be a promising paradigm for addressing critical issues in graph representation learning, such as the heavy reliance on labeled data, sparsity of data, and poor generalization. These challenges are even more significant when using LLM as the backbone architecture. Recent research highlights that the success of LLM instruction tuning is tightly dependent on factors like the volume of training data, task diversity, and annotation quality. This raises a natural question: can the principles of graph self-supervised learning be effectively transferred to settings where LLMs are the foundational architecture? To answer this, we propose two distinct types of graph self-supervised instruction data and evaluate their effectiveness in such scenarios.
\subsubsection{Self-Supervised Instructions for Topology Autoencoder} Inspired by the work of \citet{vgae}, we develop self-supervised instructions aimed at enhancing the understanding of graph topology, a fundamental capability for graph models.
The basic idea of a graph autoencoder is to encode an input graph into a latent space and reconstruct its adjacency matrix from the latent variables. We define a new learning problem, denoted by $\pi_{TAE}$, which mandates the \(f_{GFM}\) to accurately identify all direct neighbors of a query node. This task is equivalent to reconstructing the adjacency matrix, as predicting a node's first-order neighbors is both necessary and sufficient for the reconstruction. For example, we input a local social network around a user A, and the query will be \textit{``List all users with a following relationship with the user A''} and the answer should be exactly responded with all the users that meet the criteria.

\begin{center}\small \begin{tcolorbox}[title={\textcolor{gray!10}{\textbf{Topology Autoencoder Instruction Example}}}, colback=lightgray!10, colframe=gray!80, rounded corners=all, width=0.47\textwidth]
    \textbf{Input: $\boldsymbol{I_i^j \leftarrow \mathrm{Lang}_G(G_i^j,\textcolor{red}{\pi_{TAE}})}$} \\
    \text{\# Graph Description}\\
    \textcolor{gray}{$\cdots$} \\
    \text{\# Graph Text}\\
    \text{\textcolor{gray}{$\cdots$}}\\ 
    \textcolor{red}{\text{\# Query}}\\
    \text{\textcolor{gray}{\textit{List all users with a following relationship with the first user}}}  \\
    \textbf{Output: $\boldsymbol{O_i^j \leftarrow \mathrm{Lang}_Y(Y_i^j,\textcolor{red}{\pi_{TAE}})}$}\\
    \textcolor{red}{\text{\# Anwser}}\\
    \text{\textcolor{gray}{\textit{The users are ... }}}
\end{tcolorbox}\end{center}

\subsubsection{Self-Supervised Instructions for Feature Masked Autoencoder} 

Motivated by~\citet{mae2}, which demonstrates that reconstructing masked node features as the only pretext task in graph self-supervised learning could generate promising performance, we reformulate it within current framework. 

The original masked graph autoencoder first uniformly samples a subset of nodes without replacement and mask their feature with learnable embedding, and then reconstructs the masked node features from the corrupted node features and the adjacency matrix. 

Similarly, we define a learning problem $\pi_{FMAE}$. We replace the feature descriptions of a sampled subset of nodes in the graph text with text \textit{``unknown''}. Then, we require the $f_{GFM}$ to infer the raw feature descriptions based on the corrupted graph text. For example, while the text piece about user A in the corrupted input social graph will be \textit{``the hobbies of the user A are unknown''} and the response should be exactly the hobbies of the user A in the raw graph. 
 
\begin{center}\small \begin{tcolorbox}[title={\textcolor{gray!10}{\textbf{Feature Masked Autoencoder Example}}}, colback=lightgray!10, colframe=gray!80, rounded corners=all, width=0.47\textwidth]
    \textbf{Input: $\boldsymbol{I_i^j \leftarrow \mathrm{Lang}_G(\textcolor{orange}{\tilde{G}_i^j},\textcolor{red}{\pi_{FMAE}})}$} \\
    \text{\# Graph Description}\\
    \textcolor{gray}{$\cdots$} \\
    \textcolor{orange}{\text{\# Graph Text}}\\
    \textcolor{gray}{$\cdots$} \\
    \text{\textcolor{gray}{the hobbies of the user A are \sout{basketball and football} \textcolor{orange}{unknown}...}}\\ 
    \textcolor{gray}{$\cdots$} \\
    \textcolor{red}{\text{\# Query}}\\
    \text{\textcolor{gray}{\textit{Infer the missing hobbies of the user A...}}}  \\
    \textbf{Output: $\boldsymbol{O_i^j \leftarrow \mathrm{Lang}_Y(Y_i^j,\textcolor{red}{\pi_{FMAE}})}$}\\
    \textcolor{red}{\text{\# Anwser}}\\
    \text{\textcolor{gray}{\textit{The hobbies of the user A may include...}}}
\end{tcolorbox}\end{center}

In this work, we augment each input graph sample with an additional topology autoencoder sample. For input graphs containing node or edge attributes, we perturb the graph by randomly replacing 20\% of these attributes with ``unknown'' and randomly selecting a single node or edge to serve as a masked feature autoencoder sample. In Section~\ref{sec:expr_ssl}, our empirical results validate the efficacy of these two self-supervised tasks.

\begin{table*}[h]
\centering
\small
\caption{Comparison of model performance on the \benchname.. ``-'' means the model is not applicable on the task.}
\label{tab:exp-main}
\setlength{\tabcolsep}{2pt}
\renewcommand{\arraystretch}{1.0}
\begin{tabular}{cc|c|c|c|c|c|c|c|c|c|c}
\toprule
\multirow{2}{*}{Task Type} & \multirow{2}{*}{Dataset} & \multirow{2}{*}{Metric} & \multicolumn{2}{c|}{Closed} & \multicolumn{2}{c|}{Open} & \multicolumn{3}{c|}{Primitive GFM} & \multicolumn{2}{c}{Ours} \\
\cmidrule{4-12}
                                         &    &                                       & GPT-4o-mini & Qwen-plus & Llama 3 & Qwen 2 & GraphWiz & OFA & LLaGA & \modelname-I & \modelname-J \\
\midrule                                                                               
\multicolumn{1}{c}{\multirow{5}{*}{Entity-Based}} & GraphSize-Node            & Acc   & 0.925 & \bestcell{1.000} & 0.115 & 0.590  & 0.925 & -    & -    & 0.900   &  \secondcell{0.995}  \\
\multicolumn{1}{c}{}                              & GraphSize-Edge            & Acc   & 0.020 & 0.025 & 0.010 & 0.010  & 0.015 & -    & -    & \bestcell{0.325}  &  \secondcell{0.075} \\
\multicolumn{1}{c}{}                              & AttributeRetrivel-Node    & Acc   & 0.990 &  0.960& 0.515 & 0.605  & 0.080 & -    & -    & \bestcell{1.000}     &  \secondcell{0.995} \\
\multicolumn{1}{c}{}                              & AttributeRetrivel-Edge    & Acc   & 0.980 & \secondcell{0.990 }& 0.595 & 0.250  & 0.380 & -    & -    & 0.810  &  \bestcell{1.000} \\
\multicolumn{1}{c}{}                              & NodeDegree                & Acc   & 0.275 & 0.290 & 0.105 & 0.055  & 0.010 & -    & -    & \bestcell{0.510}  & \bestcell{0.510}\\
\midrule
\multirow{3}{*}{Path-Based}                       & ShortestPath              & Acc   & 0.080 & 0.140 & 0.050  & 0.010 & 0.020 & -    & -    & \bestcell{0.305}  &  \secondcell{0.215}\\
                                                  & MaximumTriangleSum        & Acc   & 0.010 & 0.090 & 0.070  & 0.005 & 0.000 & -    & -    & \secondcell{0.120}  &  \bestcell{0.130}\\
                                                  & HamiltonPath              & Acc   & \bestcell{0.621} & 0.470 & 0.394  & 0.560 & 0.449 & -    & -    & \secondcell{0.620}  &  0.591\\
\midrule
\multirow{3}{*}{Structure-Based}                  & SubgraphMatching          & Acc   & 0.490 & 0.375& 0.425   & 0.495 & 0.440 & -    & -    & \bestcell{0.630}  & \secondcell{0.510} \\
                                                  & GraphStructure            & Acc   & 0.155 & 0.130 & 0.110  & 0.185 & 0.080 & -    & -    & \bestcell{0.960}  &  \secondcell{0.955}\\
                                                  & GraphAutomorphism         & Acc   & 0.245 & 0.600 & 0.600  & 0.735 & 0.185 & -    & -    & \secondcell{0.840}  &  \bestcell{0.865}\\
\midrule
\multirow{5}{*}{Node}                             & OgbnArxiv                 & Acc   & 0.490 & \secondcell{0.650} & 0.190  & 0.340 & -     & 0.250& 0.425    & 0.595  &  \bestcell{0.660}\\
                                                  & WikiCS                    & Acc   & 0.635 & 0.695 & 0.335  & 0.370 & -     & 0.600& 0.775    & \bestcell{0.795}   &  \bestcell{0.795}\\
                                                  & Twitch                    & Acc   & 0.440 & 0.000 & 0.160  & 0.085 & -     & 0.505& 0.535    & \bestcell{0.550}  &  \secondcell{0.535}\\
                                                  & AMiner                    & Acc   & 0.660 & 0.500 & 0.240  & 0.405 & -     & 0.495& 0.280    & \secondcell{0.670}  & \bestcell{0.755}\\
                                                  & USAAirport                & RMSE  &2.159  & 1.390 & 1.411  & 1.509 & -     & -    &    -     & \bestcell{0.803}  & \secondcell{0.809}\\
\midrule
\multirow{3}{*}{Link}                             & FB15K237                  & Acc   & 0.750 & \secondcell{0.760} & 0.445  & 0.605 & -     & 0.260& 0.630   & 0.650 &  \bestcell{0.875}\\
                                                  & OgblVessel                & Acc   & 0.475 & 0.540 & 0.485  & 0.490 & -     & 0.545& 0.390 & \secondcell{0.640}  &  \bestcell{0.690}\\
                                                  & MovieLens                 & RMSE  &1.225  & \bestcell{1.143} & 1.287  & 1.204 & -     &-     &-      &\secondcell{1.173}   &  1.201\\
\midrule 
\multirow{3}{*}{Graph}                            & Fingerprint               & Acc   & 0.220 & 0.005 & 0.195  & 0.020 & -     &  -   & -    & \secondcell{0.570}  &  \bestcell{0.660}\\
                                                  & BACE                      & Acc   & 0.440 & 0.335 & 0.205  & 0.445 & -     & 0.545& -    & \secondcell{0.570}  &  \bestcell{0.585}\\
                                                  & ESOL                      & RMSE  & 2.454& 2.270  & 2.643  & 2.473 & -     & -    & -    & \secondcell{1.893}  & \bestcell{1.346}\\
\midrule
\multirow{4}{*}{Open}                             & Twitter Social Circle     & ROUGE & 0.229 & 0.186 & 0.101  &  0.150 & -    & -    & -    & \secondcell{0.492}& \bestcell{0.504} \\
                                                  & Yelp Review Generation    & ROUGE & 0.135 & 0.142 & 0.077  &0.135   & -    & -    & -    & \secondcell{0.146} & \bestcell{0.152}  \\
                                                  & Molecule Description      & ROUGE & 0.154 & 0.145 & 0.134  &0.138   & -    & -    & -    & \secondcell{0.264}& \bestcell{0.351} \\
                                                  & Cypher Query Generation   & ROUGE & 0.410 & 0.371 & 0.176  &0.448   & -    & -    & -    & \secondcell{0.556}&  \bestcell{0.749}\\
\bottomrule
\end{tabular}
\end{table*}

\section{Experiments}\label{sec:experiment}
In this section, we evaluate and analyze the performance of~\modelname~as a GFM on \benchname, as well as address the claims and designs from the methodology Section~\ref{sec:method}. Moreover, we also include a zero-shot transfer learning experiment on datasets outside of \benchname~to further validate the potential of \modelname.

In particular, we will answer the following research questions: 
\textbf{RQ1:} How effective is \modelname~as a graph model and as a GFM?
\textbf{RQ2}: Do we really need to design new textualization formats for graphs? Are the existing standard graph
exchange formats effective? 
\textbf{RQ3:} In the context of learning graph tasks in language space, is data augmentation at the text level effective? 
\textbf{RQ4:} Is traditional self-supervised learning in the graph domain still effective in the language space?
\textbf{RQ5:} How is \modelname's zero-shot transfer capability as a GFM?

\subsection{Experimental Settings}

\modelname~is based on Llama 3-8B-Instruct and employs the rank-stabilized Low-Rank
Adapters~\citep{rslora} technique for parameter-efficient fine-tuning and utilizes the RoPE scaling~\citep{ropescaling} for long context understanding. We denote \modelname-I for training on a single dataset and \modelname-J for joint training across the entire \benchname.

As for baselines, we categorize our comparisons into three groups: 
\textbf{(1) Closed-source LLMs}, including GPT-4o-mini and Qwen-plus. These models represent the highest level of reasoning ability for complex tasks and can serve as robust baselines while validating the extent of graph reasoning capability in LLMs. \textbf{(2) Open-source LLMs}, including Llama 3-8B-Instruct and Qwen-7B-Instruct. These models are comparable in size to \modelname, which can demonstrate the effectiveness of \modelname. \textbf{(3) Primitive GFMs}, including GraphWiz~\citep{chen2024graphwiz}, GraphGPT~\citep{tang_GraphGPTGraphInstruction_2024}, ZeroG~\citep{zerog}, OFA~\citep{iclr2024ofa}, and LLaGA~\citep{icml2024llaga}. These models are currently the closest to GFM and can handle a certain subset of different types of tasks or datasets with a single model. OFA is the only work capable of handling tasks at the node, edge, and graph levels simultaneously, but it is limited to classification tasks; GraphWiz is representative of various graph structure tasks; LLaGA and GraphGPT integrate graph tokenization within LLMs and primarily tackle node and edge tasks; ZeroG focuses on the zero-shot transfer capability for node classification tasks. For GraphWiz, we used the officially provided model checkpoints for inference on structural tasks. For OFA and LLaGA, we trained the official code on the proposed benchmark. For the zero-shot experiments, we utilized the best results reported in the respective papers for OFA, LLaGA, GraphGPT, and ZeroG.

\subsection{Overall Performance of \modelname~(RQ1)}
As shown in Table~\ref{tab:exp-main}, \modelname~demonstrates comparable or superior performance across nearly all tasks, whether trained independently or jointly. Notably, LangGraph excels in graph-level tasks that necessitate a deep understanding of graph structure, showcasing the effectiveness of learning graphs entirely in the language space. Analyzing the baselines can provide us more insights into \modelname's success: 

(1) Closed-source LLMs perform well on graph algorithm-related structural tasks (e.g., Hamilton Path) and knowledge graphs (i.e., FB15K237). It's known that LLMs are extensively exposed to such data during pretraining phase and common sense reasoning or math tasks in instruction tuning phase. Hence, we can deduce that the LLM's inherent complexity or capability is sufficient to model graph learning problems and serve as the backbone for GFM. We should attempt to incorporate more graph-related pretraining or instruction tuning data to stimulate the graph reasoning ability of LLMs. By expanding the model’s exposure to graph tasks, we can potentially unlock more sophisticated reasoning skills within the graph domain. 

(2) In our experiments, GraphWiz shows weak generalization on unseen data, underperforming in some tasks (e.g., Hamilton Path), aligning with its claimed tendency to overfit due to excessive instruction tuning and preference alignment on small datasets. This underscores the importance of diverse graph data and tasks in developing \modelname.

(3) In text-driven tasks like node classification or link prediction with rich text features like Ogbn-Arxiv, current quasi-GFMs seem not to have a clear advantage against LLMs. This suggests that their success in such tasks may be attributed more to how LLM encodes texts, which is often overlooked in these works since they only included traditional GNNs as baselines. 
LangGFM achieves the promising results while being able to handle all tasks simultaneously, which undoubtedly reinforces the initial question: \textit{Is there a genuine necessity for specialized modules (e.g., GNNs) in graph processing?} Perhaps we should explore more possibilities on the path to GFM.

\subsection{Graph Texuliazation Effectiveness (RQ2)}\label{sec:format_expr}
To assess the necessity of developing new custom graph representation formats, we compared the accuracy of using different formats on the Shortest Path and Ogbn-Arxiv tasks under the zero-shot in-context learning setup with Qwen-plus.  
  
\begin{figure}[ht]
 \centering
 \includegraphics[width=\linewidth]{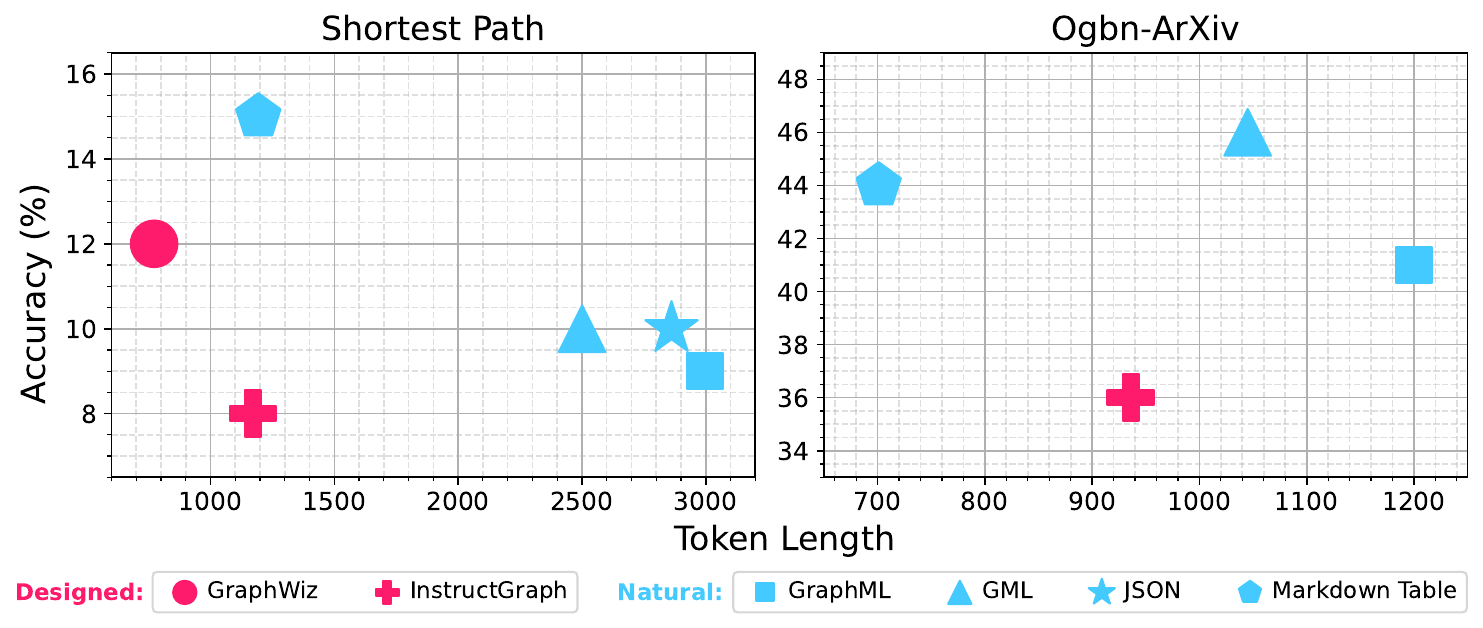}
 \caption{Preference of LLM for different graph formats.}
 \label{fig:format_pref}
\end{figure}

We used the formats proposed from GraphWiz and InstructGraph as baselines. As shown in Figure~\ref{fig:format_pref}, existing formats consistently outperformed the custom-designed ones. 
InstructGraph did not surpass any established format in accuracy, and although it demonstrated higher token efficiency on graphs without node features, this advantage disappeared when node features were introduced. GraphWiz, while offering a balanced trade-off between performance and token efficiency on featureless graphs, was not applicable for tasks involving features. 
It's observed that the Markdown Table has a relatively optimal performance-token ratio. A possible reason is the prevalence of Markdown data in code repositories (e.g., README files), where tables are often accompanied by contextual analysis, allowing the LLMs to develop a good understanding ability.  
  
In conclusion, this experiment supports the rationality of utilizing universal and widely-adopted standard graph exchange formats.

\subsection{Augmentation in Language Space (RQ3)}\label{sec:format_aug}

As discussed in Section~\ref{sec:method_format_aug}, employing diverse formats to represent the same graph structure appears to be a promising approach for graph augmentation within the language space. This strategy facilitates the development of genuine graph comprehension in LLMs by mitigating reliance on language-specific patterns and fostering a more robust understanding of the underlying graph semantics.

\begin{figure}[ht]
    \centering
    \begin{subfigure}{\linewidth}
        \centering
        \includegraphics[width=\linewidth]{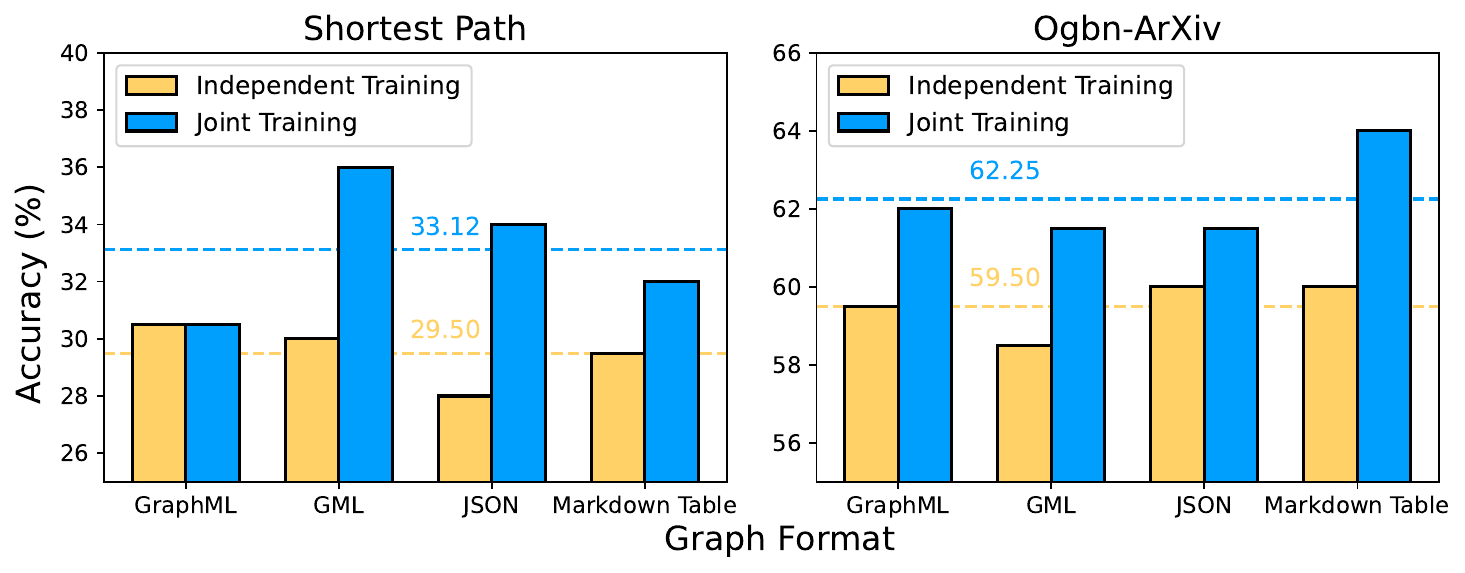}
        \vspace{-2em}
        \caption{F ormat augmentation leads to better performance.}
        \label{fig:format_aug}
    \end{subfigure}
    \begin{subfigure}{\linewidth}
        \centering
        \includegraphics[width=\linewidth]{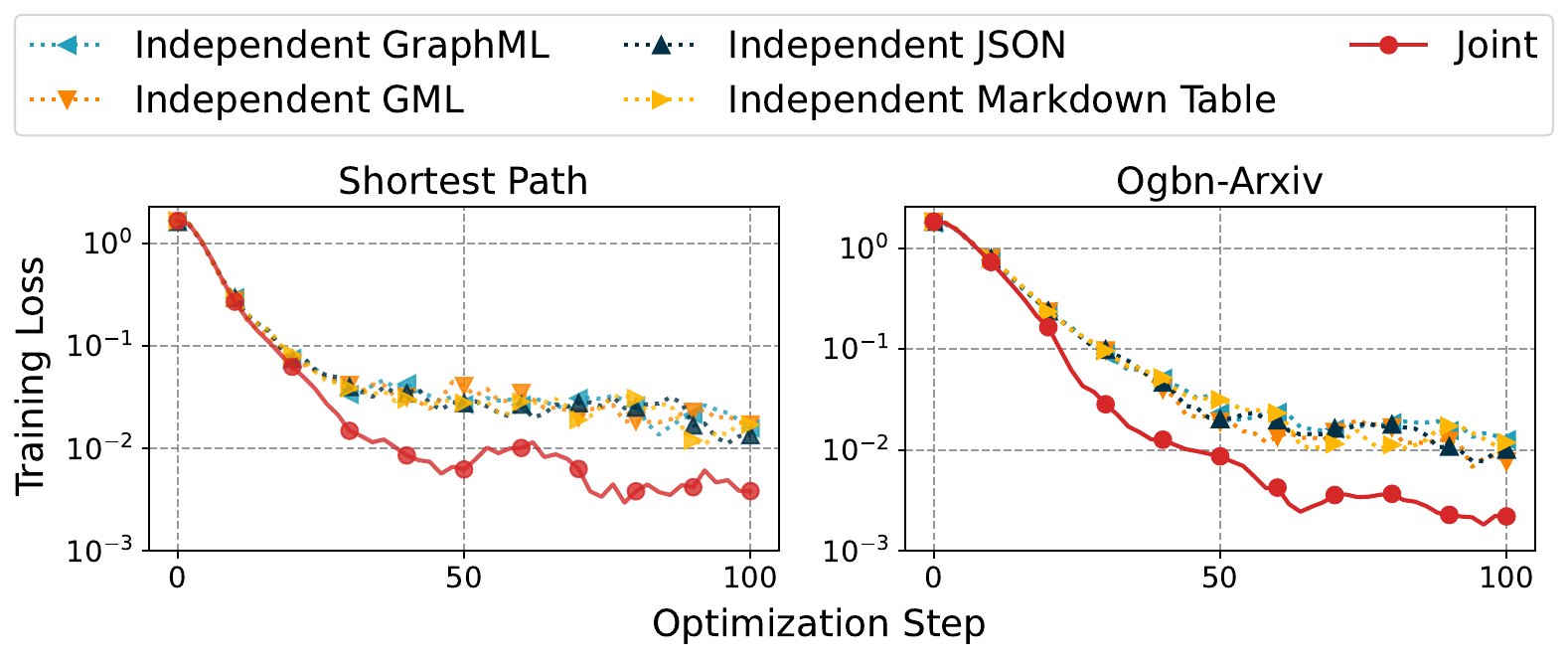}
        \vspace{-2em}
        \caption{All-format joint training converges faster and better.}
        \label{fig:format_loss}
    \end{subfigure}
    \caption{Various Formats as Graph Augmentations.}
    \label{fig:combined_figures}
\end{figure}

We conducted experiments on the Shortest Path and Ogbn-Arxiv tasks, training separately on different formats and jointly on all formats. The experimental results are shown in the figure~\ref{fig:format_aug}. 
Training on multiple formats together significantly and stably improved the task performance, essentially enhancing the model's reasoning ability in any format. Specifically, in the Shortest Path task, overall performance increased by 3.63\%, with peak performance rising from 30.5 (GraphML) to 36 (GML). For the Ogbn-Arxiv task, overall performance improved by 2.75\%, with the best score increasing from 60.0 (JSON/MARKDOWN) to 64 (Table).

Additionally, we analyzed the training loss curves for separate and joint training, as shown in Figure~\ref{fig:format_loss}. The joint training across all formats converges faster and better, further supporting the efficacy of using diverse formats as data augmentation for graph learning in the language space.

\subsection{Graph Self-supervised Learnig Effect (RQ4)}\label{sec:expr_ssl}
\begin{figure}[ht]
 \centering
 \includegraphics[width=\linewidth]{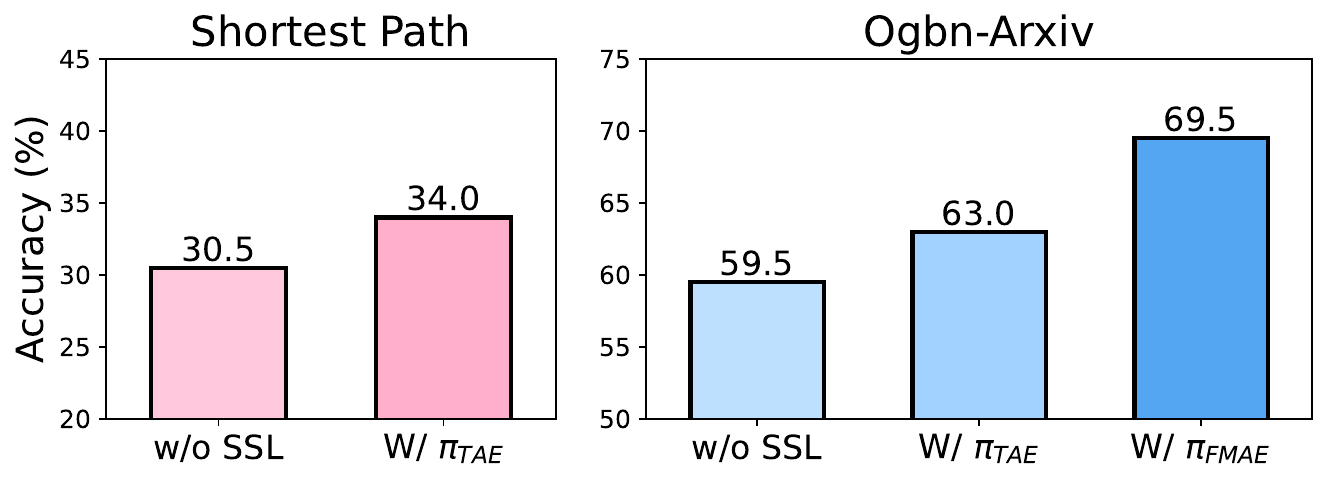}
 \caption{Effect of the proposed self-supervised learning.}
 \label{fig:ssl}
\end{figure}

Inspired by the successful experience in graph self-supervised representation learning, we designed self-supervised instructions in Section~\ref{sec:method_ssl}. $\pi_{TAE}$ represents the topology autoencoder, requiring \modelname~to correctly understand the connectivity relationships in the input graph; and $\pi_{FMAE}$ represents the feature masked autoencoder, requiring \modelname~to infer the masked node or edge features from the masked input graph. 
To address whether our designs can successfully transfer insights from conventional graph learning in learning in language space, we tested the effectiveness of $\pi_{TAE}$ on the Shortest Path task (which has no feature so that $\pi_{FMAE}$ is not applicable) and the effectiveness of both $\pi_{TAE}$ and $\pi_{FMAE}$ on the Ogbn-Arxiv task. 
The results in Figure~\ref{fig:ssl} show that $\pi_{TAE}$ can effectively improve the model's ability, increasing the accuracy of the shortest path task from 30.5 to 34 and the accuracy of Ogbn-Arxiv from 59.5 to 63. Notably, $\pi_{FMAE}$ has a particularly significant improvement on the real-world graph, increasing the accuracy of Ogbn-Arxiv from 59.5 to 69.5. 
  
These positive results demonstrate the effectiveness of the designed self-supervised learning instructions. \modelname~can better utilize the information in the graph for reasoning by better understanding the graph structure and the mechanism of graph formation. In the future, we are probably further enhance the performance of \modelname by integratinhg more graph self-supervised tasks.

\subsection{Zero-shot Transfer Learning (RQ5)}
One important capability of the foundation model is its ability for zero-shot in-context learning. For ease of comparison, we selected two commonly used zero-shot transfer learning datasets, Cora and PubMed, as well as the FreeSolv dataset from the field of molecular property prediction. The results are shown in Table~\ref{tab:exp-zeroshot}. 
Overall, our model significantly outperforms existing related GFM works. On the rich-text Cora and PubMed datasets, other baselines are notably weaker than the large models, while our model achieved stable and comparable results. Notably, on the FreeSolv dataset, our model even reached results comparable to those of domain-specific large models in molecular property prediction. These two points strongly indicate that our model possesses a deeper understanding and reasoning capability of graphs compared to current GFMs, leading to improved zero-shot transfer ability.

\begin{table}[htbp]
\centering
\small
\caption{Zero-shot transfer learning experiment.}
\label{tab:exp-zeroshot}
\begin{tabular}{ccccc}
\toprule
\multicolumn{2}{c}{\multirow{2}{*}{Methods}}& Cora  & PubMed & FreeSolv \\
                         &                   & Acc  & Acc & RMSE \\
\midrule
\multirow{2}{*}{Closed-LLMs} & Qwen Plus   & 0.625 & \bestcell{0.915}  & 32.039  \\
                             & GPT-4o-mini & \bestcell{0.705} & 0.795  & 14.997 \\
\midrule
\multirow{5}{*}{Quasi-GFM}    & OFA         & 0.231 & 0.466  & -        \\
                             & ZeroG       & 0.625 & 0.791  & -        \\
   & GraphGPT    & 0.249 & 0.399  & -        \\
                             & LLaGA       & 0.596 &   -    & -        \\
                             & MolecularGPT & - & -  & \bestcell{4.975}        \\
\midrule
   Ours   & \modelname  & \secondcell{0.635} & \secondcell{0.800}  &  \secondcell{5.491}   \\
\bottomrule
\end{tabular}
\end{table}



\section{Conclusion}
In conclusion, GFM mark a significant step forward in graph machine learning, aiming to unify complex data and enable knowledge transfer across diverse graph domains. This work introduces a comprehensive benchmark and proposes a GFM built entirely on LLM. By synthesizing techniques from both text and graph learning, we identify key limitations in current GFMs and report promising preliminary results. Our analysis highlights the potential for future advancements through improved LLM backbone and an expanded range of datasets and tasks. However, these extensions are beyond the scope of this work due to constraints in time and resources. We encourage the graph learning community to engage in further exploration of these areas. As LLMs evolve with increased context lengths and reduced training costs, the insights provided here will become increasingly impactful.

\bibliographystyle{ACM-Reference-Format}
\bibliography{custom}
















\end{document}